%% file: main.tex
\pdfoutput=1

\documentclass[11pt]{article}

\usepackage{acl}

\usepackage{times}
\usepackage{latexsym}
\usepackage[T1]{fontenc}
\usepackage[utf8]{inputenc}
\usepackage{microtype}
\usepackage{inconsolata}
\usepackage{graphicx}
\usepackage{amsmath}
\usepackage{booktabs}
\usepackage{float}

\title{Winning Big with Small Models: \\Knowledge Distillation vs.\ Self-Training\\for Reducing Hallucination in Product QA Agents}

\author{
 \textbf{Ashley Lewis\textsuperscript{1}}
 \textbf{Michael White\textsuperscript{1}}
 \textbf{Jing Liu\textsuperscript{2}}
 \\\textbf{Toshiaki Koike-Akino\textsuperscript{2}}
 \textbf{Kieran Parsons\textsuperscript{2}}
 \textbf{Ye Wang\textsuperscript{2}} \\
 \textsuperscript{1}The Ohio State University,  
 \textsuperscript{2}Mitsubishi Electric Research Laboratories \\
 \small{
 \{\href{mailto:lewis.2799@osu.edu}{lewis.2799}, 
 \href{mailto:white.1240@osu.edu}{white.1240}\}@osu.edu,  
 \{\href{mailto:jiliu@merl.com}{jiliu}, 
 \href{mailto:koike@merl.com}{koike}, 
 \href{mailto:parsons@merl.com}{parsons}, 
 \href{mailto:yewang@merl.com}{yewang}\}@merl.com}
}

\begin{document}
\maketitle
\begin{abstract}

The deployment of Large Language Models (LLMs) in customer support is constrained by hallucination—generating false information—and the high cost of proprietary models. To address these challenges, we propose a retrieval-augmented question-answering (QA) pipeline and explore how to balance human input and automation. Using a dataset of questions about a Samsung Smart TV user manual, we demonstrate that synthetic data generated by LLMs outperforms crowdsourced data in reducing hallucination in finetuned models. We also compare self-training (fine-tuning models on their own outputs) and knowledge distillation (fine-tuning on stronger models' outputs, e.g., GPT-4o), and find that self-training achieves comparable hallucination reduction.  We conjecture that this surprising finding can be attributed to increased exposure bias issues in the knowledge distillation case and support this conjecture with post hoc analysis. We also improve robustness to unanswerable questions and retrieval failures with contextualized “I don’t know” responses. These findings show that scalable, cost-efficient QA systems can be built using synthetic data and self-training with open-source models, reducing reliance on proprietary tools or costly human annotations.

\footnote{This work was conducted while Ashley Lewis was interning at Mitsubishi Electric Research Laboratories.}

\end{abstract}

\input{sections/introduction}

\input{sections/related_work}
\input{sections/data_and_experimental_setup}

\input{sections/results_and_analysis}
\input{sections/discussion}
\input{sections/conclusion}
\input{sections/limitations}
\input{sections/ethics}

\bibliography{anthology, custom}

\appendix

\input{sections/appendices/data_preprocessing}
\input{sections/appendices/question_examples}
\input{sections/appendices/prompts}
\input{sections/appendices/factscore}
\input{sections/appendices/human_eval_tutorial}
\input{sections/appendices/error_category_examples}
\input{sections/appendices/human_eval_breakdown}
\input{sections/appendices/bertscore}

\end{document}

%% file: sections/introduction.tex
\section{Introduction}
\label{sec:introduction}

\begin{figure}
    \centering
    \includegraphics[width=0.45\textwidth]{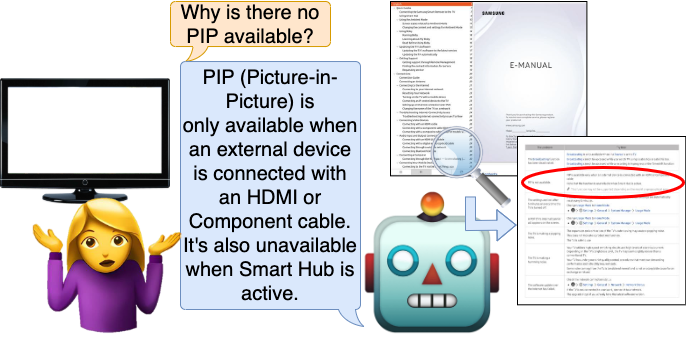}
    \caption{Overview of the retrieval-augmented QA process. A user asks a question about a product feature and the system uses relevant information from the product manual to generates a factual response.}
    \label{fig:RAG_overview}
\end{figure}

While many companies are eager to integrate Large Language Models (LLMs) into customer service and other applications, widespread deployment remains constrained by hallucination, or the generation of false or unsupported information, and the high financial and computational costs of using proprietary models. This issue is particularly critical in customer support, where unreliable responses can mislead users and erode trust.

We develop a cost-effective retrieval-augmented question-answering (QA) pipeline (see Figure \ref{fig:RAG_overview}) and address critical training data questions: what sources of data are most effective for finetuning open source models, and what preprocessing or filtering mechanisms best mitigate hallucination. To do so, we use a dataset from \citet{nandy-etal-2021-question-answering} comprising crowdsourced questions written by professional annotators about a Samsung Smart TV user manual (but notably lacking human-written responses). In this work, we address the following research questions:

\paragraph{RQ1: What is the optimal balance between manual and automated methods for data processing and creation?} We explore the trade-offs of using automatic and manual methods in two main situations: data processing and data creation. 

We use Llama-3-8B-Instruct (hereafter Llama-3) \cite{dubey2024llama3herdmodels} to generate answers to the crowdsourced questions, followed by two cleaning methods: manual cleaning performed by the first author and automatic cleaning using LLMs. While many recent studies have shown LLM's ability to iteratively evaluate and refine text to reduce hallucination \cite{dhuliawala-etal-2024-chain, wang2024selftaughtevaluators}, these methods are often costly and pose data privacy risks when proprietary models are used at runtime. To address this, we compare the effort of manual cleaning with the effectiveness of closed-source (GPT-4o) and open-source (Llama-3) models for data cleaning. We show that while GPT-4o significantly outperforms Llama-3 in cleaning quality, it is comparable to manual efforts, suggesting that manual input may not always be necessary.

We also explore a realistic scenario in which no training data is available. Perhaps surprisingly, we demonstrate that LLM-generated synthetic training data leads to lower hallucination rates than crowdsourced data, as measured by FactScore and human evaluation, possibly due to increased variability in human-written questions.

\paragraph{RQ2: How does self-training compare to model distillation in terms of hallucination rates?} We examine the benefits of synthetic data by comparing two training approaches: finetuning models on data generated by the same model (self-training with Llama-3) versus finetuning models on data generated by a stronger model (knowledge distillation using GPT-4o). \citet{lewis-white-2023-mitigating} suggest that knowledge distillation reduces hallucination, but their study only tests on synthetic questions. Meanwhile, \citet{zhang2024selfalignment} and \citet{flame} show that self-training can reduce hallucination, though without any human evaluation and with a train/test time mismatch in the case of \citet{flame}. To our knowledge, our work is the first apples-to-apples comparison of these two approaches. Surprisingly, we find that self-training of a small model and distillation of a large one achieve comparably low hallucination rates, as measured by FactScore \cite{min-etal-2023-factscore} and human evaluation, when the same data cleaning is used for both methods.

To explore this result, we analyze the potential role of exposure bias, which refers to the tendency of a model to perform better in contexts observed during training, leading to errors when faced with unfamiliar contexts during inference. We hypothesize that models trained on their own generated data benefit from greater familiarity with the training examples, compensating for the quality gap between the models. This suggests that self-training can serve as a resource-efficient alternative to model distillation in tasks where minimizing hallucination is critical.

\paragraph{RQ3: How can retrieval failures and unanswerable questions be anticipated?} The dataset includes questions scraped from community forums such as Amazon product QA sections, which are noisier, more diverse, and often unanswerable using the user manual. Such questions are prone to hallucination as the model relies on pretraining rather than the provided document. Since state-of-the-art retrieval models return n-best lists with imperfect accuracy \cite{gao2023retrieval}, it is critical for QA systems to recognize retrieval failures and respond appropriately (e.g., \textit{I don't know the answer}) while confirming the user's question was understood. While we do not focus on retrieval, we mitigate this issue by inserting negative examples during training, teaching models to provide contextualized “I don’t know” responses, which also reduces hallucination rates.

In light of these questions, this paper makes the following key contributions, with a focus on customer support systems:

\begin{itemize}
    \item We find that manual and automatic data cleaning result in finetuned models with similar factual accuracy, but responses from models based on automatic cleaning are longer.
    
    \item We demonstrate that LLM-generated synthetic training data can lead to models with lower hallucination rates than using crowdsourced data, as measured by FactScore and human evaluation.

    \item We show that finetuning a model on its own generated answers (e.g., training Llama-3 on Llama-generated data) results in comparable hallucination mitigation to training it on GPT-4o-generated answers, despite GPT-4o being a generally more capable model.

    \item We explore exposure bias as a possible explanation for why the self-trained model performs so well. We hypothesize that models perform better when trained on low-perplexity (more familiar) examples. Our FactScore results and perplexity-based analysis provide empirical support for this hypothesis.

    \item We provide a simple, scalable data perturbation strategy and synthesize contextualized \textit{I don't know} responses to increase model robustness to unanswerable questions and retrieval failures.
\end{itemize}

%% file: sections/related_work.tex
\section{Related Work}
\label{related_work}

Recent studies suggest that finetuning on new, unfamiliar knowledge can lead to hallucination \cite{gekhman-etal-2024-fine, flame, kang2024unfamiliarfinetuningexamplescontrol}.  For instance, \citet{flame} propose training on self-generated data to reduce hallucination, but introduce a training-test mismatch where models use grounding documents during training but not testing, potentially causing hallucinations. We maintain consistent setups.

Like \citet{flame}, \citet{zhang2024selfalignment} employ self-training to reduce hallucinations. Our approach differs in three ways: first, we use simple supervised finetuning (SFT) instead of techniques like reinforcement learning (RL) and direct preference optimization (DPO), which are promising avenues for future work. Second, we compare self-training with knowledge distillation, investigating the value of synthetic data from a model's own outputs and from a more performant model. Third, we validate our results with human evaluation in addition to automatic metrics. Other works also focus on iterative self-refinement \cite{wang2024selftaughtevaluators, madaan2024self}, though do not specifically focus on the problem of hallucination.

In contrast, \citet{lewis-white-2023-mitigating} employ knowledge distillation to reduce hallucination, using ChatGPT to generate and clean document-grounded training data. However, their approach is limited in two ways: they finetune a T5-large model \cite{raffel2020exploring}, which reduces hallucination over GPT-3.5 but limits robustness and fluency, and they evaluate only on synthetic data.

\citet{farquhar2024semantic} detect hallucinations during inference using semantic entropy, which clusters generated outputs based on semantic equivalence and measures uncertainty at the level of meaning.  While semantic entropy excels at runtime detection in open-domain settings, the entailment-based clustering method is very expensive.  By contrast, our approach reduces hallucinations at their source by improving training processes for RAG settings.

%% file: sections/data_and_experimental_setup.tex
\section{Data and Experimental Setup}
\label{sec:data}

\subsection{Datasets}

The primary dataset consists of 684 crowdsourced questions paired with retrieved passages from the manual \cite{nandy-etal-2021-question-answering}. We split the dataset into 534 training, 100 development, and 50 test questions (our ``regular test set''). Dataset preprocessing details can be found in Appendix \ref{app:data_preprocessing}. We focused on this dataset because many existing QA datasets either lack grounding documents or prioritize open-domain QA, which does not align with the controlled, retrieval-augmented QA setting we aimed to study. This approach also allowed us to conduct a deep-dive analysis into the trade-offs between self-training, knowledge distillation, and synthetic data generation in mitigating hallucinations within a well-defined context.

As mentioned, the dataset also contains a collection of 3,000 questions sourced from community forums. We create challenge sets by randomly selecting 100 development and 100 test questions from this set. These questions are noisier and less than half are answerable, which allows us to evaluate how well models handle particularly challenging cases. Examples from both types of questions can be found in Appendix \ref{app:example_questions}.

\subsection{Training Data}

\input{tables/test_set_factscores}

\paragraph{Regular Training Data}

We use the pretrained Llama-3-8B-Instruct \cite{dubey2024llama3herdmodels} to generate answers for the 534 training questions. Three datasets are created:
(1) a manually cleaned version where responses were reviewed and corrected by the first author, and
(2)--(3) automatically cleaned versions using GPT-4o and Llama-3-70B, respectively. This allows a systematic evaluation of the trade-offs between human effort and automated cleaning. As shown in Table \ref{tab:factscore_test_set}, cleaning with Llama-3 was largely unsuccessful. Thus in the remaining experiments GPT-4o was used for the cleaning task. We anticipate that improvements in open-source models like Llama-3 may reduce reliance on proprietary alternatives in the future. Prompts for both data generation and cleaning can be found in Appendix \ref{app:prompts}.

\paragraph{Synthetic Data} 

In addition to crowdsourced training questions, we generate fully synthetic QA data using LLMs. Specifically, we prompt Llama-3 and GPT-4o to generate new QA pairs based on passages from the Samsung Smart TV manual. To ensure that these datasets have comparable information coverage to the crowdsourced dataset and to prevent retrieval quality from being a confounding factor, we select passages systematically rather than randomly. We identify all 208 unique sections in the manual that are referenced in the crowdsourced training data. From these passages, we generate two synthetic QA pairs per passage, two from Llama-3 and two from GPT-4o. This approach ensures that the synthetic datasets are no larger than the crowdsourced dataset and cover similar content while maintaining consistency in passage selection. In a real-world application, this limitation does not exist, as synthetic training data can be generated from any number of passages. Thus, coverage is not inherently a bottleneck when using synthetic data in practical settings.

\subsection{Baseline and Experimental Models}

To evaluate the impact of data cleaning type and synthetic training data on hallucination reduction, we experiment with both pretrained models and finetuned models trained on different datasets.

\paragraph{Baseline Models}  
\begin{itemize}
    \item \textbf{Pretrained Llama-3-8B-Instruct (Llama-3)}: An open-source model that serves as a strong starting point for retrieval-augmented generation (RAG) without task-specific adaptation \cite{dubey2024llama3herdmodels}. The model is run with few-shot prompting.
    \item \textbf{GPT-4o}: A state-of-the-art proprietary model, included as a benchmark to assess how well finetuned open-source models compare to a highly optimized general-purpose system \cite{openai2024gpt4technicalreport}. The model is run with few-shot prompting.
\end{itemize}

\paragraph{Finetuned Models}  

We finetune Llama-3 on different variations of training data to analyze the effects of data source, cleaning method, and exposure bias on hallucination rates. Specifically, we train models on the following datasets using supervised fine-tuning (SFT) with LoRA adapters, following the parameters and framework of \citet{zheng-etal-2024-llamafactory}. During inference, we use greedy decoding with default settings:

\begin{itemize}
    \item \textbf{Manually Cleaned Training Data}: A dataset where the first author reviewed and corrected Llama-3-generated answers to the \citet{nandy-etal-2021-question-answering} 534 crowdsourced training questions.
    \item \textbf{Automatically Cleaned Training Data}: A version of the training set where errors in Llama-3-generated answers were identified and repaired using GPT-4o.
    \item \textbf{Synthetic Data (Llama vs.\ GPT)}: Two datasets where  416 QA pairs were generated by either Llama-3 or GPT-4o based on passages from the Samsung Smart TV manual. All synthetic data was cleaned using GPT-4o.
    \item \textbf{Synth Llama+}: Trained on the synthetic Llama data, and augmented with 100 negative examples (see section \ref{synth_llama+} for more details).
\end{itemize}

\subsection{Metrics for Evaluation}

We evaluate model performance using two methods: FactScore \cite{min-etal-2023-factscore}, an automated metric for factual accuracy, and human evaluation by trained annotators. These complementary approaches measure factual consistency and response quality.

\paragraph{FactScore}

FactScore evaluates whether a model's response aligns with a reference document. It works by decomposing a response into sentences, breaking each sentence into discrete factual claims, and verifying their alignment with the reference text. FactScore measures the proportion of supported claims while penalizing hallucinated content. However, responses from GPT-4o and SynthGPT, which often use structured formatting (e.g., lists, topic headers), cause FactScore to produce fragmented or nonsensical claims, unfairly penalizing these models. To address this, we removed the sentence-splitting preprocessing and instead generated atomic facts directly from the full response.

FactScore, which we computed using GPT-4o-mini, has been shown to be a reliable proxy for factuality, correlating well with human judgments \cite{min-etal-2023-factscore}. However, we find that it is unsuitable for evaluating \textit{I don’t know} responses. Thus, we applied FactScore only to the regular test set (mostly answerable questions), excluding the challenge set (many unanswerable questions). We also used it to evaluate human-written training questions for synthetic models, as they do not see these at training time and it provides a more robust evaluation. Further information in Appendix \ref{app:factscore}.

\input{tables/annotation_criteria}

\paragraph{Human Evaluation} 
To obtain a more nuanced assessment of response quality, we conducted a human evaluation with three fluent English speaking, Linguistics PhD students (instructions in Appendix \ref{app:human_eval}), who annotated each model-generated response for the regular test set (50 items) and 50 items from the challenge set. They assigned to each response one of the categories listed in Table \ref{tab:error_categories} (examples in Appendix \ref{app:error_category_examples}), which were determined by an author analysis of the dev set. Three-way agreement occurred between annotators 63.14\% of the time and two-way agreement occurred 36.43\% of the time. Krippendorff's Alpha was $\alpha$ = 0.625, indicating substantial agreement.

Each response was labeled independently by all three annotators. The final assigned label was determined by a majority vote. In the few cases where annotators provided three different labels, the response was assigned the most severe error based on the following predefined ranking:  Hallucination > Non-Answer > Partial Answer > IDK - Bad > Disfluent > Other. The purpose of this ranking is to prioritize hallucination and content errors. For example, if a response is labeled as ``Hallucination,'' ``Good,'' and ``Partial Answer,'' it is assigned the final label of ``Hallucination'' due to its higher severity in the ranking.

By combining automated and human evaluation, we ensure a comprehensive analysis of both quality and factual consistency in model-generated responses. The aggregated results can be found in Table \ref{tab:humaneval} and the separate results on the regular and challenge test sets can be found in Appendix \ref{app:human_eval_breakdown}.

\input{tables/lengths}

%% file: tables/test_set_factscores.tex
\begin{table}[t]
    \centering
    \renewcommand{\arraystretch}{1.2}
    \setlength{\tabcolsep}{6pt}
    \begin{tabular}{l|c}
        \toprule
        \textbf{Model} & \textbf{FactScore} \\
        \midrule
        Llama-3      & 0.9077 \\
        GPT-4o         & 0.9323 \\
        \hline
        Uncleaned       & 0.8798 \\
        Manual cleaned  & 0.8810 \\
        Autocleaned\textsubscript{L} & 0.8202 \\
        Autocleaned\textsubscript{G}     & 0.8966 \\
        \hline
        SynthGPT   & 0.9116 \\
        SynthLlama & 0.9211 \\
        SynthLlama+ & \textbf{0.9461} \\
        \bottomrule
    \end{tabular}
    \caption{FactScore results for the test set. Pretrained base models: Llama-3 and GPT-4o. Finetuned Llama-3-8B models on the \citet{nandy-etal-2021-question-answering} dataset: Uncleaned (no data cleaning performed), Manual cleaned (cleaning done by the first author), Autocleaned\textsubscript{L} and Autocleaned\textsubscript{G} (cleaning done by Llama-3-70B and GPT-4o, respectively). Finetuned Llama-3-B models on synthetic data: SynthGPT (trained on data generated by GPT-4o), SynthLlama (trained on data generated by Llama-3-8B), and SynthLlama+ (same as SynthLlama, with additional negative examples).}
    \label{tab:factscore_test_set}
\end{table}

%% file: tables/annotation_criteria.tex
\begin{table}[t]
    \centering
    \renewcommand{\arraystretch}{1.3}
    \begin{tabular}{p{2cm}p{4.75cm}}
        \toprule
        \textbf{Category} & \textbf{Description} \\
        \midrule
        \textbf{Hallucination} & The response contains information not present in the manual. \\
        \textbf{Non-Answer} & The response does not answer the question. \\
        \textbf{Partial answer} & The response does not fully answer the question, or omits important information. \\
        \textbf{IDK - Bad} & The manual section has the information required to answer the question, but the response is mistakenly ``I don’t know''. \\
        \textbf{Disfluent} & The response contains grammatical or fluency problems. \\
        \textbf{Other} & The response contains some other type of error. \\
        \textbf{IDK - Good} & The manual section does not contain the information required to answer the question and the response is appropriately ``I don’t know''. \\
        \textbf{Good} & There are no errors. \\

        \bottomrule
    \end{tabular}
    \caption{Response error categories and their descriptions. Examples can be found in Appendix \ref{app:error_category_examples}.}
    \label{tab:error_categories}
\end{table}

%% file: tables/lengths.tex
\begin{table}[t]
    \centering
    \small 
    \renewcommand{\arraystretch}{1.2} 
    \setlength{\tabcolsep}{4pt}
    \begin{tabular}{lccc}
        \toprule
        \textbf{Model} & \textbf{Chall. (100)} & \textbf{Reg. (50)} & \textbf{Total (150)} \\
        \midrule
        Pretrain & 26.56 & 28.74 & 27.29 \\
        GPT-4o & 22.23 & 31.56 & 25.34 \\
        \hline
        Manual & 21.74 & 28.54 & 24.01 \\
        Auto-cleaned & 26.33 & 31.00 & 27.89 \\
        \hline
        SynthLlama & 36.06 & 44.56 & 38.89 \\
        SynthGPT & 40.40 & 47.34 & 42.71 \\
        SynthLlama+ & 21.92 & 42.06 & 28.63 \\
        \bottomrule
    \end{tabular}
    \caption{Average response lengths for different models across challenge and regular test sets.}
    \label{tab:lengths}
\end{table}

%% file: sections/results_and_analysis.tex
\section{Results and Analysis}
\label{results_and_analysis}

\input{tables/human_evaluation}

\subsection{Autocleaning vs. Manual Cleaning}

The FactScore results on the test set (Table \ref{tab:factscore_test_set}) and human evaluation results (Table \ref{tab:humaneval}) reveal that models finetuned on autocleaned data perform slightly better in terms of factual accuracy and response quality compared to manually cleaned data, though the gains are small. No models were significantly better than pretrained Llama-3.
  
Table~\ref{tab:lengths} shows that responses generated from the model trained on autocleaned data are consistently longer than those from manually cleaned data, suggesting that autocleaning prioritizes including as much information as possible from the retrieved passage, even when it is unnecessary to answer the question. This verbosity, while occasionally useful, does not inherently improve factuality.

The response quality of autocleaned and manually cleaned models is similar, as indicated by FactScore and human evaluation results. Both outperform a model trained on uncleaned data but fail to surpass the pretrained Llama-3 baseline. However, hallucination remains a persistent issue across all models, regardless of the cleaning method.

One reason for the lack of significant improvements between manual and autocleaned models may be the limited training data (only 534 examples), which likely reduces the relative impact of cleaning strategies. Furthermore, the absence of sufficient negative training examples, such as explicit ``I don’t know'' responses, leaves models prone to over-generating information rather than admitting uncertainty—an issue particularly evident in the challenge test set.

Importantly, while the cleaning strategies evaluated here do not independently outperform the pretrained baseline, their primary utility lies elsewhere: enabling the generation of higher-quality synthetic QA data. As described in Section \ref{synth_data_section}, models finetuned on synthetic data derived from cleaned examples (e.g., SynthLlama, SynthGPT) significantly outperform both manually and automatically cleaned models. This suggests that cleaning should be viewed not as an end in itself, but as a preparatory step for creating effective training data in low-resource settings.

\subsection{Human vs. Synthetic Training Data}
\label{synth_data_section}

\input{tables/diversity_metrics}

A key question in this study is whether crowdsourced training data is necessary for finetuning QA models, or if synthetically generated data can achieve comparable or even superior performance. We compare models trained on crowdsourced answers against those trained on LLM-generated synthetic data (from Llama-3 and GPT-4o), evaluating them on both the regular and challenge test sets. 

Table \ref{tab:factscore_test_set} and Table \ref{tab:humaneval} indicate  that models trained on synthetic data can outperform those trained on crowdsourced data in terms of factual accuracy and overall response quality. One possible explanation is that crowdsourced data tend to introduce variability and noise, whereas synthetic data is consistently aligned with the retrieved passages and the LLM's internal language patterns, making it easier for the model to learn structured answer generation. 

In Table \ref{tab:diversity} we examine diversity using GEM metrics \cite{gehrmann-etal-2021-gem} and find that crowdsourced questions, while shorter on average, have a larger vocabulary of distinct 1-, 2-, and 3-grams relative to the number of total tokens, suggesting greater diversity. We also calculate BERTScores \cite{zhang2020bertscore} for every pair of questions within each dataset and find that, on average, the scores for the synthetic data are higher, indicating that the questions are more semantically similar to each other than the questions in the crowdsourced dataset. We also calculate the perplexity of the questions for Llama-3 and find higher perplexity in the human questions, indicating that they are more unfamiliar to the model. While greater diversity can potentially be helpful in finetuning a model, evidently the less diverse and more expected synthetic questions are more consistently helpful in our experiments. Further analysis can be found in Appendix \ref{app:bertscores}.

\subsection{Synth Llama+: Enhancing Synthetic Data for Hallucination Reduction}
\label{synth_llama+}

To encourage the model to abstain from answering when relevant information is unavailable, as is often the case in the challenge test set, we added negative training examples to the synthetic Llama data by duplicating 100 random training questions. Then, instead of generic ``I don’t know'' responses, we constructed context-aware refusals by replacing the correct passage with a random one and prompting Llama-3 to generate an answer using these items. This ensured that the model could acknowledge the user’s intent while signaling retrieval failure, as shown in the following example:

\begin{quote}
\textbf{Question:} How do I select Dynamic mode?

\textbf{Passage:} The compression of video content may cause picture distortions, especially in fast-moving pictures from sports programs and action movies. [...]

\textbf{Generated Response:} I'm sorry, I can't find any information about selecting Dynamic mode in the provided section of the user manual.
\end{quote}

Unlike generic refusals, this approach ensures that the model’s response acknowledges the intent of the question, making it clear to users that their request was understood but that relevant information is unavailable. We select SynthLlama here because it provides the best balance of low cost and high performance, which is an important consideration for real-world applications.

These enhancements led to improvements in both FactScore and human evaluation metrics compared to the base SynthLlama model and comparable performance to GPT-4o on this task.  With these improvements, SynthLlama+ achieved a significantly higher proportion of good responses in comparison to pretrained Llama in the human evaluation, as shown in Table~\ref{tab:humaneval}.

\subsection{Exposure Bias and Synthetic Data Performance}

\input{tables/train_set_factscores}

\begin{figure}
    \centering
    \includegraphics[width=0.3\textwidth]{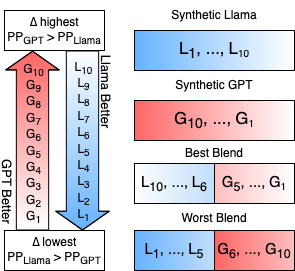}
    \caption{A toy example of 10 training items per synthetic model to demonstrate how the Best and Worst 50:50 blends were created.}
    \label{fig:data_blending}
\end{figure}

One of the key findings in our study is that self-trained models perform comparably to knowledge-distilled ones---that is, models finetuned on synthetic data generated by the same model (e.g., Llama-3 trained on Llama-generated QA pairs) perform about as well as those trained on synthetic data from a more performant model (e.g., Llama-3 trained on GPT-generated QA pairs) when both synthetic datasets use data cleaning. This suggests that exposure bias may influence training stability and factual accuracy, as models appear to be more reliable when finetuned on data that aligns closely with their pretraining distribution. Exposure bias in language models refers to the mismatch between training and inference: during training, the model learns with gold context (``teacher forcing''), but at inference, it generates text based on its own prior predictions, potentially causing errors to accumulate and degrade output quality \cite{arora-etal-2022-exposure}. 

To further investigate this conjecture, we used the pretrained Llama-3 model to compute the perplexity of each QA response, conditioned on the passage. To quantify the relative familiarity of each synthetic example, we calculated the difference in perplexity between the GPT-generated and Llama-generated QA for each passage,

\begin{equation}
  \label{eq:difference}
\Delta PP = PP(q_{\text{G}}, a_{\text{G}} \mid c) - PP(q_{\text{L}}, a_{\text{L}} \mid c)
\end{equation}

\noindent
where \( (q_{\text{G}}, a_{\text{G}}) \) and \( (q_{\text{L}}, a_{\text{L}}) \) are the question-answer pairs generated by GPT-4o and Llama-3 for passage \( c \), respectively, and \( PP(q, a \mid c) \) represents the perplexity score of a given QA pair under the pretrained Llama-3 model.

This measure allows us to rank training examples based on their relative familiarity to the base Llama-3 model. Positive values (\( \Delta PP>0 \)) indicate that the GPT-generated QA pair is more perplexing (i.e., less familiar) to the model than the Llama-generated QA pair, whereas negative values (\( \Delta PP<0 \)) suggest the opposite.

We then sorted all passages by their perplexity difference (\(\Delta PP\)) and constructed the Best and Worst 50:50 Blends as follows. See Figure \ref{fig:data_blending} for a visual of this process using a toy example.

\paragraph{Best Blend} For each passage, we selected the QA pair where the generating model had a larger perplexity advantage relative to the other model. This means selecting the 50\% of GPT-generated QA pairs where \( \Delta PP \) is smallest and the 50\% of Llama-generated QA pairs where \( \Delta PP \) is largest.
\paragraph{Worst Blend} For each passage, we selected the QA pair where the generating model had a larger perplexity disadvantage relative to the other model. This means selecting the 50\% of GPT-generated QA pairs where \( \Delta PP \) is largest and the 50\% of Llama-generated QA pairs where \( \Delta PP \) is smallest.

Each blend contained an equal mix (50\% GPT-generated and 50\% Llama-generated), ensuring a direct comparison of training effects when models are finetuned on their most versus least familiar examples relative to each other.

\paragraph{Results and Analysis}

Table~\ref{tab:factscore_training} shows the FactScore results for the regular training set questions. Because these manually-written questions are not used at training time for the synthetic models, they can be repurposed as a larger test set, allowing for significant differences to emerge. The results reveal no significant difference between synthetic GPT and synthetic Llama, suggesting comparable performance. Meanwhile, the Worst Blend model performs significantly worse than the Best Blend model, indicating that the perplexity of the training examples does play a role in the downstream model's propensity to hallucinate. Meanwhile, the Best Blend model has a higher score than both synthetic models, suggesting that perplexity-based selection could be a tool worth exploring further in mitigating hallucination for synthetic data.

%% file: tables/human_evaluation.tex
\begin{table*}[t]
    \centering
    \renewcommand{\arraystretch}{1.2}
    \resizebox{\textwidth}{!}{
    \begin{tabular}{lcccccccc|c}
        \toprule
        \textbf{Model} & \textbf{Halluc.} & \textbf{Non-Ans} & \textbf{Partial} & \textbf{IDK - Bad} & \textbf{Disfl.} & \textbf{Other} & \textbf{IDK - Good} & \textbf{Good} & \textbf{Total Good} \\
        \midrule
        Pretrained  & 13  & 0  & 6  & 0  & 1  & 5 & 24  & 51  & 75 \\
        GPT-4o  & 9   & 0  & 2  & 1  & 0  & 0 & 29  & 59  & 88 \\
        \hline
        Manual cleaned  & 14  & 2  & 7  & 0  & 3 & 5 & 21  & 48  & 69 \\
        Autocleaned\textsubscript{G}  & 13  & 0  & 6  & 0  & 2  & 9 & 19  & 51  & 70 \\
        \hline
        SynthGPT  & 9   & 0  & 0  & 2  & 3  & 8 & 22  & 56  & 78 \\
        SynthLlama  & 7   & 0  & 2  & 0  & 2 & 7 & 26  & 56  & 82 \\
        SynthLlama+ & \textbf{6}   & 0  & 0  & 0  & 1 & 2 & 31  & \textbf{60} & \textbf{  91}* \\
        \bottomrule
    \end{tabular}
    }
    \caption{Human evaluation results in which 3 annotators assessed response quality across multiple error categories for the regular test set (50 items) and 50 items from the challenge test set. Majority vote decided the final category for each item, and in cases where all 3 annotators disagreed, the most severe error is the final category. SynthLlama+ had a significantly higher proportion of good items (p < .05) over pretrained Llama, $\chi^2(1, N=100) = 9.1, p = .0026$. No other results were significant.}
    \label{tab:humaneval}
\end{table*}

%% file: tables/diversity_metrics.tex
\begin{table}[t]
    \centering
    \resizebox{\columnwidth}{!}{
    \begin{tabular}{lccc}
        \toprule
        Metric & SynthGPT & SynthLlama & Human \\
        \midrule
        Distinct-1 & 0.083 & 0.082 & 0.100 \\
        Distinct-2 & 0.263 & 0.270 & 0.345 \\
        Distinct-3 & 0.400 & 0.407 & 0.541 \\
        Mean length & 13.853 & 14.269 & 9.659 \\
        Mean perplex & 13.356 & 13.027 & 15.339 \\
        Mean BERTScore & 0.644 & 0.630 & 0.554 \\
        \bottomrule
    \end{tabular}
    }
    \caption{
        Metrics of questions from the human and synthetic datasets. 
        \textbf{distinct-1, -2, and -3} measure the proportion of unique unigrams, bigrams, and trigrams relative to the total number of tokens.
        \textbf{Mean length} refers to the average length of the questions in terms of tokens.
        \textbf{Mean perplexity} is calculated relative to Llama-3-8B.
        \textbf{Mean BERTScore} is the average of scores of every pair of questions in the dataset.}
    \label{tab:diversity}
\end{table}

%% file: tables/train_set_factscores.tex
\begin{table}[t]
    \centering
    \renewcommand{\arraystretch}{1.2}
    \setlength{\tabcolsep}{6pt}
    \begin{tabular}{l|c}
        \toprule
        \textbf{Model} & \textbf{FactScore} \\
        \midrule
        Worst Blend & 0.8826 \\
        Synthetic Llama & 0.8883 \\
        Synthetic GPT & 0.8956 \\
        Best Blend & \textbf{0.9103} \\
        \bottomrule
    \end{tabular}
    \caption{FactScore results on the training set of human-written questions. Only the Best Blend model was significantly higher than the Worst Blend model with T-Statistic 3.2858 and p-value 0.0011.}
    \label{tab:factscore_training}
\end{table}

%% file: sections/discussion.tex
\section{Discussion}
\label{discussion}

Our findings demonstrate that self-training and knowledge distillation can be comparably effective in reducing hallucination, while self-training is much less costly. Models trained on self-generated data consistently performed as well or better than those trained on GPT-generated data, supporting the hypothesis that exposure bias plays a key role in finetuning effectiveness. Additionally, our Best Blend vs.\ Worst Blend analysis revealed that using high-perplexity examples at training time led to increased hallucination, reinforcing the importance of training on familiar, low-perplexity data. Further improvements were observed with Synth Llama+, where incorporating simple, context-aware negative examples yielded higher factual accuracy, suggesting promising future directions for hallucination mitigation.

While our experiments focus on a single domain, the underlying mechanisms behind exposure bias and synthetic data effectiveness are likely to generalize to other QA tasks. Applying this approach in domains such as medical or legal QA would provide a valuable test of its robustness and effectiveness in higher-stakes applications.

Future work should explore scaling synthetic data generation, refining data selection methods based on perplexity differences, and investigating iterative self-training approaches, where models continuously refine their own synthetic data over multiple training cycles. This could further enhance model alignment and factuality while reducing reliance on external supervision.

%% file: sections/conclusion.tex
\section{Conclusion}

In this work, we explore the trade-offs between cost, manual effort, and performance in building a QA agent for customer service, with a focus on mitigating hallucination. We elucidate the components of this process that can be automated and what models are best for that automation. We find that models finetuned on synthetic datasets can outperform ones from crowdsourced datasets, and that self-training with data validation not only matches the performance of knowledge distillation but can rival the original model being distilled (GPT-4o). Our findings suggest that using this approach, scalable and cost-effective QA systems can be rapidly developed for customer service applications, delivering performance comparable to or exceeding that of current state-of-the-art models.

%% file: sections/limitations.tex
\section{Limitations}

Despite these insights, our study has limitations. First, our test set size is relatively small, particularly for human evaluation, where only 50 challenge and 50 regular test items were labeled. We did not want to overwhelm our annotators with too large of a task and judged that this was the maximum we could require. This limits the statistical power of our findings, making it difficult to detect smaller but meaningful performance differences. Expanding the evaluation set and conducting a larger-scale human evaluation in future work could provide a clearer picture of the impact of different training strategies. Our work focuses on low-resource, domain-specific QA, reflecting common real-world settings—particularly in customer support—where large annotated datasets are rarely available. To our knowledge, the SmartTV corpus we use is the only publicly available product-manual QA dataset of its kind with a permissible license.

Second, measuring hallucination remains challenging. FactScore, while useful, is not a perfect proxy for factuality, and human judgments, though more reliable, are limited by annotator agreement and scale. More robust hallucination metrics, particularly those that better capture the subtle ways in which models generate misleading but plausible responses, would enhance future analyses.

Thirdly, we limit our experiments by using only Llama-3-8B as our base model. Our primary goal was to isolate the impact of training strategies—namely, self-training versus knowledge distillation—rather than compare model families. To ensure a fair comparison, we held the base model architecture constant across experimental conditions. Llama-3-8B was selected as a strong, cost-effective, and widely adopted open-source model. This choice supports reproducibility and reflects standard practice in related work; several recent papers on hallucination mitigation (e.g., \citet{zhang2024selfalignment} and \citet{flame}) also restrict their experiments to only Llama-based models. However, future work with other architectures would be important to ensure generality of our findings here.

%% file: sections/ethics.tex
\section{Ethics}

\subsection{Data Usage and Privacy}
Our research utilizes synthetic data generated by large language models (LLMs) and publicly available and licensed datasets from user manuals for consumer electronics. All data used in this study is devoid of personally identifiable information (PII) and does not infringe upon individual privacy rights. The synthetic data generation process was carefully designed to ensure that no sensitive or identifiable information is included. Our institution's review board reviewed our human evaluation plans and ruled that it does not meet the federal definition of human subjects research requiring review. Our human evaluators were unpaid volunteer colleagues and were informed about how their annotations would be used.

\subsection{Use of Proprietary Models}
Our work leverages GPT-based models in several instances, including as comparison (baseline) models, for synthetic data generation, and in the automatic data cleaning pipeline. While GPT models are not fully reproducible due to their proprietary nature, their use in this work is limited to tasks where their high performance offers meaningful value. Specifically:

\begin{itemize}
    \item GPT is used as a baseline model to benchmark the performance of open-source systems.
    \item GPT-generated synthetic data is provided alongside the Llama-generated data to enable future reproducibility of experiments.
    \item GPT is employed for data cleaning because it demonstrates state-of-the-art performance for this specific task. The study shows that both manual and automated cleaning yield similar outcomes.
    \item To address concerns about reproducibility, all synthetic datasets and cleaned data used in the study will be made publicly available. This ensures that future researchers can reproduce our results even if proprietary models like GPT are unavailable.

\end{itemize}

Note also that GPT-4o was used as a writing assistant for this paper in a limited capacity (rephrasings, help with conciseness) and with some coding tasks during research.

\subsection{Potential Risks and Mitigation}

While our study focuses on reducing hallucinations and improving factual accuracy in QA systems, we acknowledge potential risks related to synthetic data, which may introduce subtle biases or inaccuracies. Because this domain is specific to a product user manual, we did not feel that this was a relevant issue and we did not see any problematic instances of such biases. 

\subsection{Societal Impact}

Our research aims to enhance the accuracy and reliability of QA systems, particularly in retrieving and synthesizing information from structured documents like user manuals. This can improve accessibility and user experience. However, we are aware of the broader implications of deploying such systems in real-world settings, as we demonstrate in this study that these models are still capable of hallucination even in our best-performing settings.

\subsection{Transparency and Reproducibility}
We are committed to transparency and reproducibility in our research. Despite the use of proprietary GPT-based models, our findings do not hinge on the unique capabilities of GPT. The use of GPT is supplementary and not central to the key contributions of this work. To ensure reproducibility, we will provide all synthetic datasets, cleaned data, and detailed descriptions of our experimental methodologies.

%% file: sections/appendices/data_preprocessing.tex
\section{Data Preprocessing}
\label{app:data_preprocessing}

The dataset used in this study required extensive preprocessing to align the Samsung Smart TV user manual with the accompanying QA pairs and to ensure the data was suitable for a retrieval-augmented QA framework. This process involved converting the manual into a structured format and addressing inconsistencies in the original QA dataset.

\subsection{Unused Components of the Provided Dataset}

The dataset provided by \citet{nandy-etal-2021-question-answering} includes several components for QA tasks over electronic device manuals. While we relied heavily on their crowdsourced Samsung Smart TV QA dataset, other components were excluded due to specific limitations, outlined below:

\paragraph{1. Pretraining Corpus of Product User Manuals}
This corpus, designed for pretraining, was not used due to:
(1) Formatting Issues: It contained significant noise, including garbled characters, mixed languages, and missing elements like images and titles, likely due to automated PDF-to-text conversion.
(2) Irrelevance: Pretraining on this noisy data was unnecessary, as this study focused on fine-tuning QA systems and retrieval-augmented methods.

\paragraph{2. Galaxy S10 User Manual and QA Dataset}
The Galaxy S10 manual and its associated dataset of 50 crowdsourced questions were excluded because:
(1) Subset Issues: The questions were a small subset of a larger, unreleased dataset, raising potential licensing concerns.
(2) Scale: With only 50 questions, this dataset lacked the scale required for meaningful experimentation, especially compared to the Samsung Smart TV QA dataset.

\subsection{User Manual Preparation}

The Samsung Smart TV manual, originally provided as a PDF, presented several challenges for direct use. The JSON format provided was inconsistent, likely due to automatic conversion processes, and the structure of the manual did not align well with the ``Section Hierarchy'' fields used in the QA dataset, which point to the part of the manual from which the passage is retrieved. Unfortunately, an initial search for a reliable PDF conversion tool yielded few satisfactory results. To address these issues, the first author undertook a semi-manual process to convert the manual into a structured JSON format.

First, screenshots of the original manual's table of contents were taken to map its hierarchical structure. Using GPT-4o, we generated a nested JSON representation that mirrored this hierarchy, with sections and subsections organized into dictionaries. The text within each section was carefully transcribed into corresponding fields, and images were replaced with placeholders (e.g., [image\_X.png]) that referenced a separate folder containing labeled images. To get transcriptions, we first fed each section of the manual to GPT-4o and asked it to fill in the section of the new JSON file. This was a very iterative process, with the first author manually checking the transcriptions and updating as necessary. This approach ensured that the JSON file was both faithful to the manual's structure and practical for passage retrieval tasks. Manual adjustments were made throughout the process to correct formatting errors and inconsistencies, ensuring the final structure was robust and usable.

\subsection{Cleaning the Crowdsourced QA Dataset}

The QA dataset included human-written questions linked to specific spans of text within the manual. However, the dataset required significant cleaning to align with the newly structured manual. Many questions contained incorrect ``Section Hierarchy'' fields, which were manually corrected to match the updated JSON structure of the manual.

Additionally, we expanded the retrieved passages associated with each question. Instead of limiting retrieval to short spans, we included entire sections from the manual, reflecting a more realistic retrieval scenario for QA systems. These adjustments not only improved the alignment between the questions and the manual but also made the dataset more suitable for the task of mitigating hallucinations.

\subsection{Constructing the Challenge Dataset}

Included in the \citet{nandy-etal-2021-question-answering} dataset are a collection of ~3,000 real-world user questions sourced from community forums. The questions seem to primarily come from the Amazon product pages of various Samsung Smart TVs. While there is variety in these products (model, size, etc.), they all use the same software and general hardware described in the user manual. There are many questions in this collection that are not answerable by the user manual, however. While the answers from the product pages are included, they are not reliable as (1) there is no guarantee that they are correct, (2) could involve subjective opinions, (3) may not correspond to information available in the user manual, thus we are unable to match the responses to grounding passages. Because of this, we do not rely on the answers as a resource. According to the \citet{nandy-etal-2021-question-answering} paper, there are annotations for which of these questions are answerable using the manual, but it does not seem that these annotations were publicly available.

Further, these questions do not have corresponding retrieved passages, which are necessary for our experiments. However, because these questions are only used at test and validation time and because their usefulness stems from their unanswerability, we could rely on less-than-perfect means of finding corresponding passages. Thus we simply feed the entire user manual JSON to GPT-4o and ask it to identify the most relevant passage for each of the randomly selected 100 questions in the dev and test set (200 total). This proved to be the quickest and easiest way to find passages, but a more reliable and realistic method would have been to use a state-of-the-art retrieval model. In an analysis of the dev set, we found that only 26\% of the questions are answerable. 

%% file: sections/appendices/question_examples.tex
\section{Examples of Questions from the Dataset}
\label{app:example_questions}

The following are two examples of questions from the crowdsourced dataset:

\begin{enumerate}
    \item  \textbf{Question}: How do I get better audio quality. What are the connections guidelines for it?

    \textbf{Retrieved Document}:

    For better audio quality, it is a good idea to use an AV receiver.
    
    If you connect an external audio device using an optical cable, the Sound Output setting is automatically changed to the connected device. However, to make this happen, you must turn on the external audio device before connecting the optical cable. To manually change the Sound Output setting, do one of the following:
    
    - Use the Quick Settings screen to change to the connected device: Use the Select button to select Audio Out/Optical on the Sound Output menu. ([HOME] > [SETTINGS] Settings > up directional button > Sound Output).
    
    - Use the Settings screen to change to the connected device: Select Audio Out/Optical on the Sound Output menu. ([HOME] > [SETTINGS] Settings > Sound > Sound Output).
    
    An unusual noise coming from a connected audio device while you are using it may indicate a problem with the audio device itself. If this occurs, ask for assistance from the audio device's manufacturer.
    
    Digital audio is only available with 5.1 channel broadcasts.

    \item  \textbf{Question}: How do I access the main accessibility menu to change Voice Guide settings?

    \textbf{Retrieved Document}:

    You can also go to an accessibility menu from the TV settings menu. This provides more options, for example, to change the speed of Voice Guide.
    
    The TV will not verbalize this menu unless Voice Guide is already turned on.
    
    1. Press the HOME button.
    
    2. Press the left directional button until you reach Settings.
    
    3. Press Select and a menu will open.
    
    4. Press the down directional button to reach General, and then press Select to open this menu.
    
    5. Use the directional buttons to go to the Accessibility menu, and then press Select to open this menu.
    
    6. The menu will appear with Voice Guide Settings being the first menu. Highlight Voice Guide Settings, and then press Select.
    
    7. A menu appears with the options to change Voice Guide and Volume, Speed, Pitch.
    
    8. Select the menu using the directional buttons, and then press Select.
    
\end{enumerate}

The following are two examples of questions from the challenge set (from community forums):

\begin{enumerate}
    \item \textbf{Question}: Does this tv allow me to play contents from my ipad or iphone?
    
    \textbf{Retrieved Document}:
    
    English > Connections > Connecting Your Mobile Device > Text

    You can install the SmartThings app from App Store or Google Play Store.

    \textbf{Answer}: Yes.

    \item \textbf{Question}: What is the return policy if I don't like it?

    \textbf{Retrieved Document}:
    
    English > Troubleshooting > Getting Support > Requesting service

    [HOME] > Settings > Support > Request Support
    
    You can request service when you encounter a problem with the TV. Select the item matching the problem that you encountered, and then select Request Now or Schedule Appointment > Send. Your service request will be registered. The Samsung Contact Center will contact you to set up or confirm your service appointment.

    [NOTE] You must agree to the terms and conditions for the service request.
    
    [NOTE] This function may not be supported depending on the geographical area.
    
    [NOTE] This function requires an Internet connection.

    \textbf{Answer}: You won't want to return it as it's the best in its 32 inch class.
\end{enumerate}

%% file: sections/appendices/prompts.tex
\section{Generation and Cleaning Prompts}
\label{app:prompts}

\subsection{Answer Generation Prompt}

The following is the prompt given to GPT-4o and base Llama-3-8B to generate answers to the training set questions from \citet{nandy-etal-2021-question-answering}. It uses one-shot prompting, first providing a QA example.

\begin{quote}
\ttfamily
    
Please answer the following question using the information within the section of the user manual provided. Keep the answers short and conversational.

1

***QUESTION:  

Where do I find Bixby guide?

***DOCUMENT:  

Press and hold the [MIC] button on your Samsung Smart Remote, say a command, and then release the [MIC] button. The TV recognizes the voice command.  

To view the Bixby guide, press the [MIC] button once:  

When you press the [MIC] button for the first time, the [Using Bixby] button appears at the bottom of the screen. Press the [Select] button. The [Using Bixby] popup window appears, and a tutorial on using Bixby is shown.  
When you press the [MIC] button after the first time, the [Enter My Bixby] button appears at the bottom of the screen. Press the [Select] button to go to the My Bixby screen.  

[image\_4.png]

***ANSWER:  

The Bixby guide can be found by pressing the mic button once. The first time, a `using Bixby' button will appear. Click that for setup.

2

***QUESTION: 

[TARGET QUESTION]

***DOCUMENT:  

[REFERENCE PASSAGE FOR TARGET QUESTION]

***ANSWER:  

\end{quote}

\subsection{Evaluation Prompt}

The following is the first stage of data cleaning in which GPT-4o is asked to evaluate each response and identify errors. It uses two-shot prompting.

\begin{quote}
    \ttfamily

    Your job is to evaluate the answers in the following scenarios. Given the sections of the user manual and the questions, please assess the answers and label them with one of the following categories:

    1. Good. There are no errors.
    
    2. Partial answer. The answer does not fully respond to the question, or omits important information from the manual.
    
    3. Answer not available. The manual does not contain the information required to answer the question.
    
    4. Disfluent. The answer contains grammatical mistakes or fluency problems.
    
    5. Hallucination. The answer contains information that did not come from the manual.
    
    6. Other. The answer contains some other type of error.
    
    If the label is not ``good'', please provide a short explanation.

    1
    
    QUESTION:
    
    Can I select Motion Lighting ?
    
    USER MANUAL SECTION:
    
    Reducing the energy consumption of the TV
    
    [HOME] > Settings > General > Eco Solution
    
    You can adjust the brightness level of the TV, reduce overall power consumption, and prevent overheating.
    
    Motion Lighting: Adjusts the brightness in response to on-screen movements to reduce power consumption.
    
    Auto Power Off: Automatically turns off the TV to reduce unnecessary power consumption if there is no operation for 4 hours.
    
    ANSWER:
    
    Yes, you can adjust the Motion Lighting to reduce the TV's power consumption.
    
    EVALUATION:
    
    Partial answer. The answer does not explain how to select motion lighting. It should have said that you can do so by going to [HOME]>Settings>General>Eco Solution.

    2
    
    QUESTION:
    
    What is the use of universal guide?
    
    USER MANUAL SECTION:
    
    Using the Universal Guide App
    
    Search for and enjoy content such as TV shows, dramas, movies, sports broadcasts, and music.
    
    [HOME] > [UNIVERSAL GUIDE] Universal Guide
    
    [image\_27.png]
    
    [NOTE] The image on your TV may differ from the image above depending on the model and geographical area.
    
    Universal Guide is an app that allows you to search for and enjoy various content such as TV shows, dramas, movies, and music in one place. Universal Guide can recommend content tailored to your preferences and notify you of new drama series.
    
    You can use this feature on your mobile with Samsung SmartThings app.
    
    [NOTE] To enjoy the content from these apps on your TV, they must be installed on the TV.
    
    [NOTE] When you watch some paid content, you may need to make a payment using their associated app.
    
    [NOTE] Images may look blurry depending on the service provider's circumstances.
    
    [NOTE] This function may not be supported depending on the model or geographical area.
    
    ANSWER:
    
    The universal guide allows you to search for content, like TV shows, movies, and music.
    
    EVALUATION:
    
    Good. There are no errors.

    3

    QUESTION:
    
    [TARGET QUESTION]
    
    USER MANUAL SECTION:
    
    [REFERENCE PASSAGE FOR TARGET QUESTION]
    
    ANSWER:
    
    [TARGET ANSWER]
    
    EVALUATION:

\end{quote}

\subsection{Rewrite Prompt}

The following is the prompt used to have GPT-4o rewrite incorrect answers with as few edits as possible, using one-shot prompting. It is given the question, user manual passage, answer, and the evaluation text from the previous step. Answers that are labeled ``Good'' in the previous step are skipped.

\begin{quote}
\ttfamily

    Below are question and answer pairs, each using a provided document. They may have errors, and I have provided feedback about what that error might be. Your job is to use that feedback to rewrite the answer to better answer the question and utilize the provided section of a user manual for a Samsung Smart TV. Try to make the rewrites as minimal as possible, but make whatever changes are necessary to fix the problem according to the evaluation.

    1
    
    QUESTION:
    
    Can I select Motion Lighting?
    
    USER MANUAL SECTION:
    
    Section: System and Support>Using the Screen Burn Protection and Energy Saving Functions>Reducing the energy consumption of the TV
    
    [HOME] > Settings > General > Eco Solution
    You can adjust the brightness level of the TV, reduce overall power consumption, and prevent overheating.
    Motion Lighting: Adjusts the brightness in response to on-screen movements to reduce power consumption.
    Auto Power Off: Automatically turns off the TV to reduce unnecessary power consumption if there is no operation for 4 hours.
    
    ANSWER:
    
    Yes, you can adjust the Motion Lighting to reduce the TV's power consumption.
    
    EVALUATION:
    
    Partial answer. The answer does not explain how to select motion lighting. It should have said that you can do so by going to [HOME]>Settings>General>Eco Solution.
    
    REWRITE:
    
    Yes, you can adjust the Motion Lighting to reduce the TV's power consumption by going to [HOME]>Settings>General>Eco Solution.

    2

    QUESTION:
    
    [TARGET QUESTION]
    
    USER MANUAL SECTION:
    
    [REFERENCE PASSAGE FOR TARGET QUESTION]
    
    ANSWER:
    
    [TARGET ANSWER]
    
    EVALUATION:

    [TARGET EVALUATION]

    REWRITE:
    
\end{quote}

%% file: sections/appendices/factscore.tex
\section{FactScore}
\label{app:factscore}

\subsection{Sentence Splitting}
The FactScore methodology consists of breaking generated responses first into sentences, and then takes each of those sentences and generates atomic facts from them. These atomic facts are then individually verified as True or False.

As mentioned, we found that FactScore often unfairly penalized responses that contained structured formatting (e.g., lists, topic headers) which are most common in the outputs of GPT-4o and SynthGPT. This seems to occur due to the sentence splitting not handling such formatting well. For example, this generated response from GPT:

\begin{quote}
    To get Samsung Remote Support for your TV, follow these steps:

    1. Call the Samsung Contact Center and ask for remote support.
    
    2. On your TV, go to the menu and select [HOME] > Settings > Support.
    
    3. In the Support section, select Remote Management, then read and agree to the service agreements.
    
    4.. When the PIN screen appears, provide the PIN number to the agent.
    
    5. The agent will then access your TV to provide remote support.

\end{quote}

When we calculate FactScore with the original preprocessing step (using sentence splitting), we get these atomic facts with their annotations:

\begin{table}[H]
\centering
\begin{tabular}{cp{5.5cm}c}
1 & Samsung Remote Support is available for your TV. & True \\ 
2 & There are specific steps to follow to access Samsung Remote Support. & True \\ 
3 & \textbf{The steps for accessing support are not provided in the sentence.} & False \\ 
4 & You should call the Samsung Contact Center. & True \\ 
5 & \textbf{The number for the contact center is 1.} & False \\ 
6 & You should ask for remote support. & True \\ 
7 & You should go to the menu on your TV. & True \\ 
8 & Select [HOME]. & False \\ 
\end{tabular}
\end{table}

\begin{table}[H]
\centering
\begin{tabular}{cp{5.5cm}c}
9 & Choose Settings. & False \\ 
10 & Select Support. & False \\ 
11 & There is a Support section. & True \\ 
12 & Users should select Remote Management in the Support section. & True \\ 
13 & Users need to read the service agreements. & True \\ 
14 & Users are required to agree to the service agreements. & True \\ 
15 & A PIN screen appears. & True \\ 
16 & The user should provide the PIN number to the agent. & True \\ 
17 & \textbf{The PIN number is 5.} & False \\ 
18 & The agent will access your TV. & True \\ 
19 & The purpose of accessing the TV is to provide remote support. & False \\ 
\end{tabular}
\end{table}

In this example you can see that the deconstruction of the list makes the numbers confusing to the model (facts 5 and 17) and that the model is confused by not having access to the remainder of the response in fact 3.

In contrast, without the sentence splitting, the following facts are generated from this response:

\begin{table}[H]
\centering
\begin{tabular}{cp{5.5cm}c}

1 & To get Samsung Remote Support for your TV, you need to call the Samsung Contact Center. & True \\ 
2 & You should ask for remote support when you call. & True \\ 
3 & On your TV, you need to go to the menu. & True \\ 
4 & You should select [HOME] > Settings > Support. & True \\ 
5 & In the Support section, you need to select Remote Management. & True \\ 
6 & You must read and agree to the service agreements. & True \\ 
7 & When the PIN screen appears, you need to provide the PIN number to the agent. & True \\ 
8 & The agent will access your TV to provide remote support. & True \\ 
\end{tabular}
\end{table}

As you can see, these facts are much more sensible and better reflect the content of the response.

\subsection{\textit{I Don't Know} Responses}

As mentioned, FactScore turns out to be unhelpful in assessing ``I don't know'' responses. For example, the generated response is:

\begin{quote}
    Unfortunately, the provided section does not mention turning on the TV using voice. It only provides information on turning the TV on using the [POWER] button.
\end{quote}

And the decomposed atomic facts are:

\begin{table}[H]
\centering
\begin{tabular}{cp{5.5cm}c}
1 & The provided section does not mention turning on the TV using voice. & False \\
2 & It provides information on turning the TV on using the [POWER] button. & True \\
\end{tabular}
\end{table}

The resulting FactScore for this response is .5 (quite low) despite the response being appropriate. Because of this undesired penalty, we do not use FactScore to evaluate the challenge set, as it consists of mostly \textit{I don't know} responses.

%% file: sections/appendices/human_eval_tutorial.tex
\section{Human Evaluation Tutorial}
\label{app:human_eval}

Human evaluators were instructed to review the following slide deck prior to beginning the evaluation. The slides provide instructions for how to annotate items and examples of errors (from the dev set) -- see Appendix \ref{app:error_category_examples}.

\begin{figure}[H]
    \centering
    \includegraphics[width=0.45\textwidth]{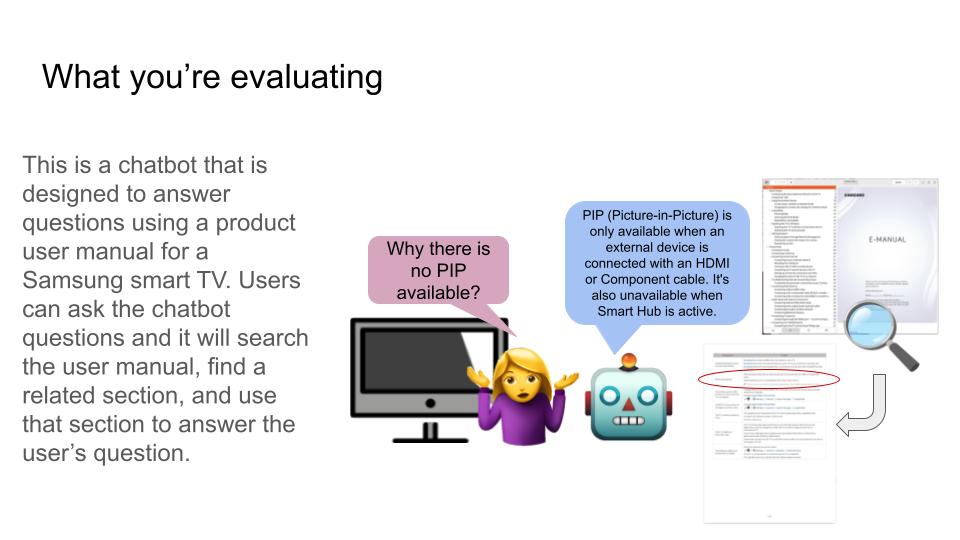}
    \caption*{Slide 1}
    \label{fig:eval2}
\end{figure}

\vspace{-1cm}

\begin{figure}[H]
    \centering
    \includegraphics[width=0.45\textwidth]{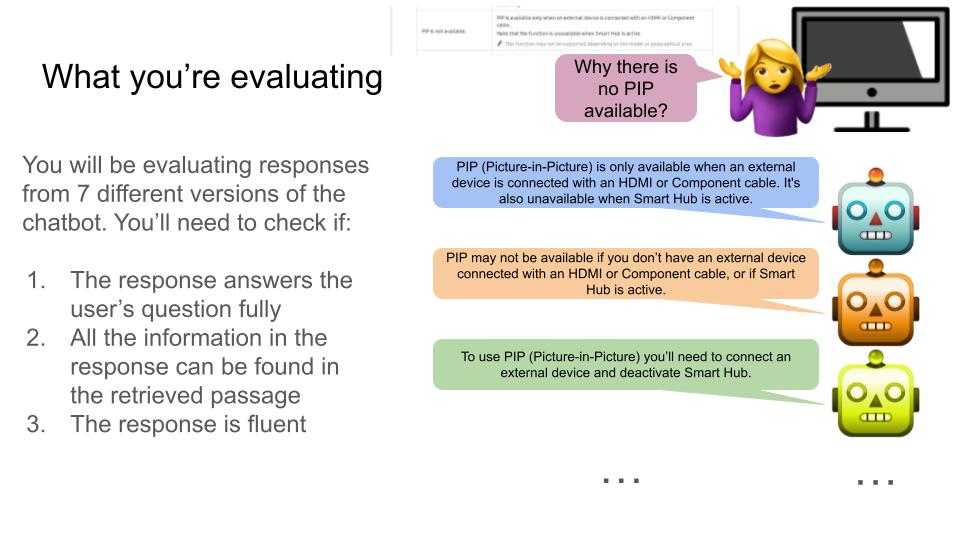}
    \caption*{Slide 2}
    \label{fig:eval3}
\end{figure}

\vspace{-1cm}

\begin{figure}[H]
    \centering
    \includegraphics[width=0.5\textwidth]{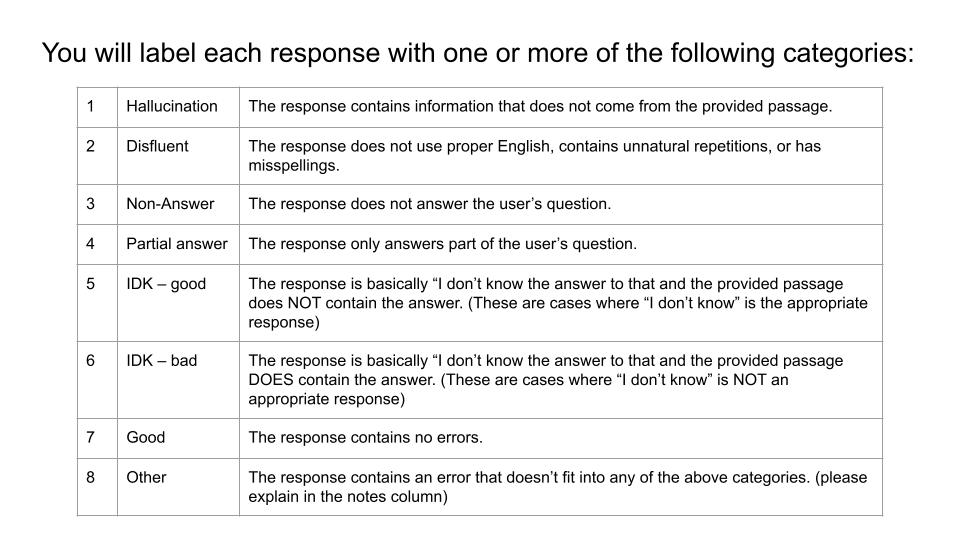}
    \caption*{Slide 3}
    \label{fig:eval4}
\end{figure}

\noindent Slides 4 - 7 show the same examples as Appendix \ref{app:error_category_examples} and thus have been omitted here.

\begin{figure}[H]
    \centering
    \includegraphics[width=0.5\textwidth]{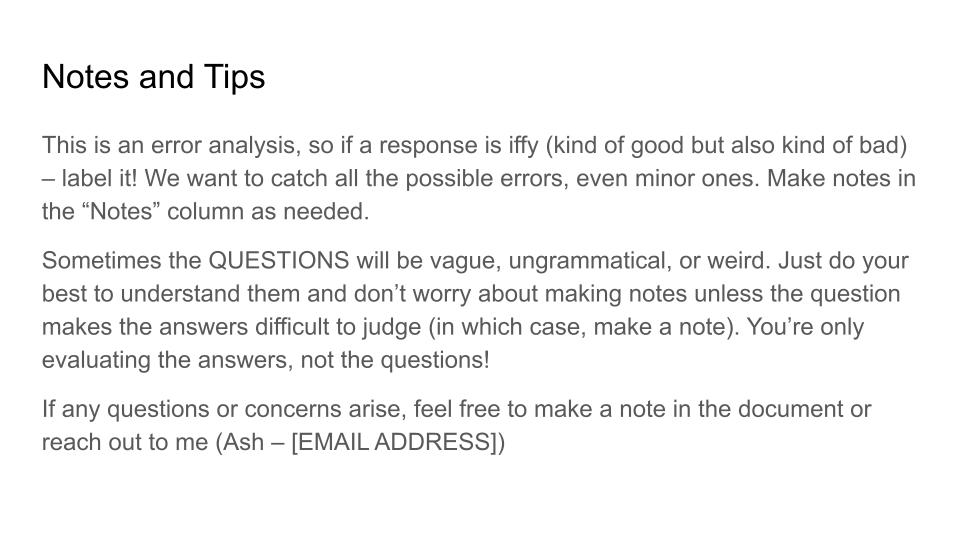}
    \caption*{Slide 8}
    \label{fig:eval9}
\end{figure}

%% file: sections/appendices/error_category_examples.tex
\section{Error Category Examples}
\label{app:error_category_examples}

The following examples come from the regular and challenge dev sets and were included in the tutorial for human evaluation in Appendix \ref{app:human_eval}. The error categories are as follows:

\begin{itemize}
    \item \textbf{Hallucination}: The response contains information not present in the manual.
    \item \textbf{Non-Answer}: The response does not answer the question.
    \item \textbf{Partial Answer}: The response does not fully answer the question, or omits important information.
    \item \textbf{IDK - Bad}: The manual section has the information required to answer the question, but the response is mistakenly ``I don’t know''.
    \item \textbf{Disfluent}: The response contains grammatical or fluency problems.
    \item \textbf{Other}: The response contains some other type of error.
    \item \textbf{IDK - Good}: The manual section does not contain the information required to answer the question and the response is appropriately ``I don’t know''.
    \item \textbf{Good}: There are no errors.
\end{itemize}

\subsection{Example Set 1}

\textbf{Question}: Why the TV smells of plastic?

\vspace{.25cm}

\noindent\textbf{Retrieved Passage}:

\noindent Troubleshooting > Other Issues > The TV smells of plastic.

\noindent This smell is normal and will dissipate over time.

\vspace{.25cm}

\noindent \textbf{Responses:}

\begin{enumerate}
    \item \textbf{Response:} The smell of plastic is normal and will dissipate over time. \\
          \textbf{Evaluation:} Good \\
          \textbf{Explanation:} The answer is correct.
    \item \textbf{Response:} If the TV smells of plastic, you should contact customer service. \\
          \textbf{Evaluation:} Hallucination \\
          \textbf{Explanation:} This response contradicts the information in the retrieved passage.  
    \item \textbf{Response:} The TV smells of plastic is normal and will go away over time. \\
          \textbf{Evaluation:} Disfluent \\
          \textbf{Explanation:} The answer is not grammatically correct.               
    \item \textbf{Response:} I’m sorry, I’m not sure of the answer to that. You can check the user manual. \\
          \textbf{Evaluation:} IDK – Bad \\
          \textbf{Explanation:} The response is “I don’t know” but the retrieved passage DOES contain the answer, so the response should have answered the question.
\end{enumerate}

\subsection{Example Set 2}

\textbf{Question}: how many inches does it stick out off the wall?

\vspace{.25cm}

\noindent\textbf{Retrieved Passage}:

\noindent English > Precautions and Notes > Read After Installing the TV > Installing an Anti-theft Lock

\noindent An anti-theft lock is a physical device that can be used to protect the TV against theft. Look for the lock slot on the back of the TV. The slot has a [LOCK] icon next to it. To use the lock, wrap the lock cable around an object that is too heavy to carry and then thread it through the TV's lock slot. The lock is sold separately. The method of using an anti-theft lock may differ for each TV model. Refer to the lock's user manual for more information.

\noindent [NOTE] This function may not be supported depending on the model or geographical area.

\vspace{.25cm}

\noindent \textbf{Responses:}

\begin{enumerate}

    \item \textbf{Response:} The provided document does not mention the inches of protrusion off the wall. It only discusses the anti-theft lock installation. \\
          \textbf{Evaluation:}  IDK – Good \\
          \textbf{Explanation:} The answer cannot be found in the retrieved passage, so the response is appropriate.

\end{enumerate}

\subsection{Example Set 3}

\textbf{Question}: Does this tv have an antenna output and does it have the vesa hole pattern for wall mounting?

\vspace{.25cm}

\noindent\textbf{Retrieved Passage}:

\noindent English > Connections > Connecting an Antenna

\noindent You can connect an antenna cable to your TV.

\noindent [NOTE] An antenna connection is not necessary if you connect a cable box or satellite box.

\noindent [image\_7.png]

\vspace{.25cm}

\noindent \textbf{Responses:}

\begin{enumerate}
    \item \textbf{Response:} Yes, the TV has an antenna output, and you can connect an antenna cable to it. Additionally, the TV has a VESA hole pattern for wall mounting. \\
          \textbf{Evaluation:} Hallucination \\
          \textbf{Explanation:} While the answer to the antenna question is correct, the response makes up information about the VESA hole pattern.

    \item \textbf{Response:} Yes, this TV has an antenna output. \\
          \textbf{Evaluation:} Partial answer \\
          \textbf{Explanation:} The answer only contains a response to the first part of the question.

    \item \textbf{Response:} Yes, this TV has an antenna output, but the user manual passage doesn’t mention the VESA hole pattern for wall mounting, so I cannot answer that question. \\
          \textbf{Evaluation:} IDK – good \\
          \textbf{Explanation:} Even though the answer to the first question is correct, it still contains an “IDK” response to the second question
          
\end{enumerate}

\subsection{Example Set 4}

\textbf{Question}: Is this tv hdr 10?

\vspace{.25cm}

\noindent\textbf{Retrieved Passage}:

\noindent English > Quick Guides > Connecting the Samsung Smart Remote to the TV

\noindent Connect the Samsung Smart Remote to your TV to operate the TV.
When you turn on the TV for the first time, the Samsung Smart Remote pairs to the TV automatically. If the Samsung Smart Remote does not pair to the TV automatically, point it at the remote control sensor of the TV, and then press and hold the [GO BACK] and [PLAY] buttons simultaneously for 3 seconds or more.

\noindent [image\_1.png]

\noindent [NOTE] The images, buttons, and functions of the Samsung Smart Remote may differ depending on the model.

\noindent [NOTE] For more information about the Samsung Smart Remote that comes with the QLED TV, refer to "About the Samsung Smart Remote (QLED TV)."

\noindent [NOTE] For more information about the Samsung Smart Remote that comes with the UHD TV, refer to "About the Samsung Smart Remote (UHD TV)."

\noindent [NOTE] The Samsung Smart Remote may not be supported depending on the model or geographical area.

\vspace{.25cm}

\noindent \textbf{Responses:}

\begin{enumerate}
    \item \textbf{Response:} No, this document does not mention HDR 10 or any other HDR format. \\
          \textbf{Evaluation:}  Other \\
          \textbf{Explanation}: This response has an “other” error because it correctly says “IDK” but it starts the response with “No, …” which makes it seem like the answer to the user’s question is no, which is misleading.
\end{enumerate}

%% file: sections/appendices/human_eval_breakdown.tex
\clearpage

\onecolumn

\section{Human Evaluation: Regular and Challenge Test Sets}
\label{app:human_eval_breakdown}

\subsection{Regular Test Set Results}

\begin{table}[H]
    \centering
    \renewcommand{\arraystretch}{1.2} 
    \resizebox{\textwidth}{!}{
    \begin{tabular}{lcccccccc|c}
        \toprule
        \textbf{Model} & \textbf{Halluc.} & \textbf{Non-Ans} & \textbf{Partial} & \textbf{IDK - Bad} & \textbf{Disfl.} & \textbf{Other} & \textbf{IDK - Good} & \textbf{Good} & \textbf{Total Good} \\
        \midrule
        Pretrained  & 4  & 0  & 4  & 0  & 0  & 0 & 1  & 41  & 42 \\
        GPT-4o  & 2   & 0  & 1  & 0  & 0  & 0 & 1  & 46  & 47 \\
        \hline
        Manual  & 4  & 0  & 5  & 0  & 1 & 0 & 1  & 39  & 40 \\
        Autocleaned\textsubscript{G}  & 4  & 0  & 4  & 0  & 2  & 0 & 0  & 40  & 40 \\
        \hline
        SynthGPT  & 2  & 0  & 0  & 0  & 2  & 0 & 1  & 45  & 46 \\
        SynthLlama  & 2  & 0  & 1  & 0  & 1  & 0 & 1  & 45  & 46 \\
        SynthLlama+  & 2  & 0  & 0  & 0  & 1  & 0 & 1  & 46  & 47 \\
        \bottomrule
    \end{tabular}
    }
    \caption{Human evaluation results on the Regular Test set, assessing response quality across various error categories. Majority vote determined the final category for each item.}
    \label{tab:humaneval_updated_2}
\end{table}

\subsection{Challenge Test Set Results}

\begin{table}[H]
    \centering
    \renewcommand{\arraystretch}{1.2} 
    \resizebox{\textwidth}{!}{ 
    \begin{tabular}{lcccccccc|c}
        \toprule
        \textbf{Model} & \textbf{Halluc.} & \textbf{Non-Ans} & \textbf{Partial} & \textbf{IDK - Bad} & \textbf{Disfl.} & \textbf{Other} & \textbf{IDK - Good} & \textbf{Good} & \textbf{Total Good} \\
        \midrule
        Pretrained  & 9  & 0  & 2  & 0  & 1  & 5 & 23  & 10  & 33 \\
        GPT-4o  & 7   & 0  & 1  & 1  & 0  & 0 & 28  & 13  & 41 \\
        \hline
        Manual  & 10  & 2  & 2  & 0  & 2 & 5 & 20  & 9  & 29 \\
        Autocleaned\textsubscript{G}   & 9  & 0  & 2  & 0  & 0  & 9 & 19  & 11  & 30 \\
        \hline
        SynthGPT  & 7  & 1  & 0  & 2  & 1  & 8  & 21  & 11  & 32 \\
        SynthLlama  & 5  & 0  & 1  & 0  & 1  & 7  & 25  & 11  & 36 \\
        SynthLlama+  & 4  & 0  & 0  & 0  & 0  & 2  & 30  & 14  & 44 \\
        \bottomrule
    \end{tabular}
    }
    \caption{Human evaluation results on the Challenge Test Set, assessing response quality across various error categories. Majority vote decided the final category for each item.}
    \label{tab:humaneval_updated}
\end{table}
\twocolumn

%% file: sections/appendices/bertscore.tex
\section{Human vs. Synthetic Data Analysis}
\label{app:bertscores}

In order to get a better sense of the differences between the datasets, we plot the distribution of BERTScores for each. As you can see, the human-written questions cluster lower, meaning that fewer questions are very similar to each other. Both sets of synthetic questions cluster higher and more evenly, suggesting less variety.

\begin{figure}[H]
    \centering
    \includegraphics[width=0.4\textwidth]{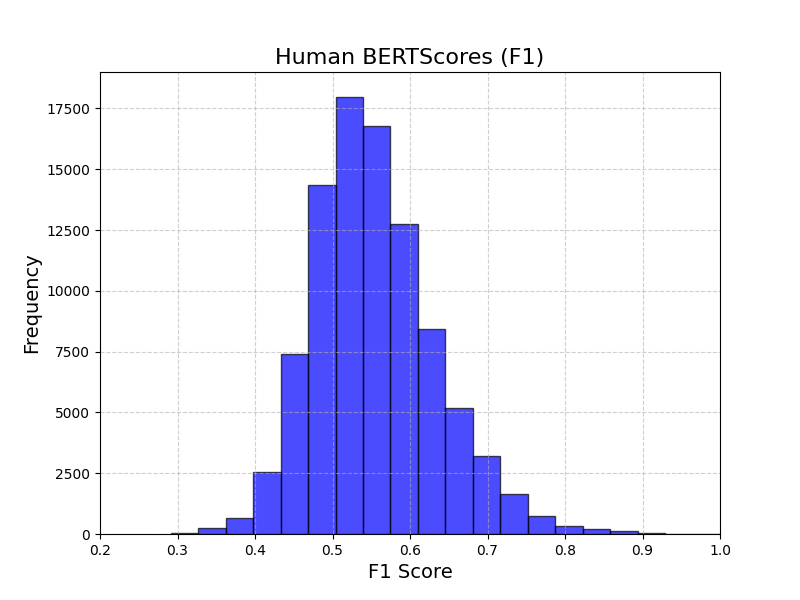}
    \caption{Distribution of the BERTScores for every combination of two questions in the crowdsourced dataset.}
    \includegraphics[width=0.4\textwidth]{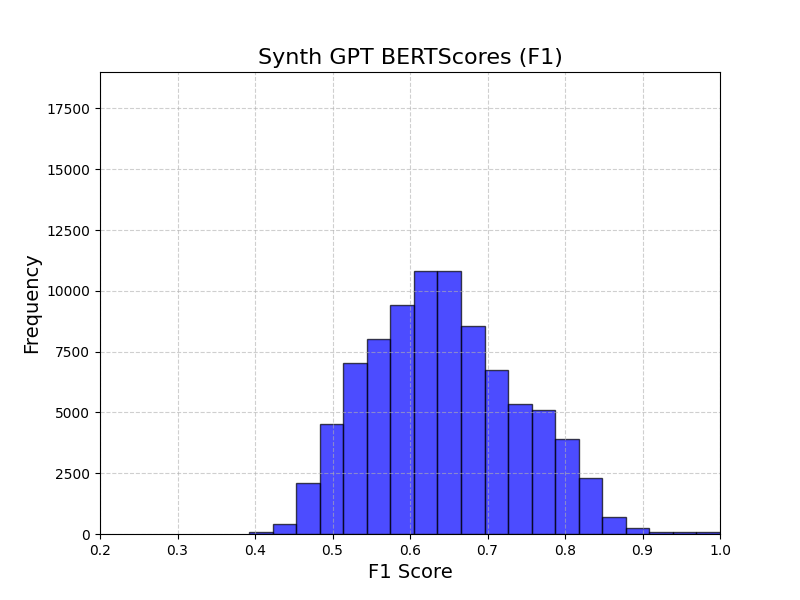}
    \caption{Distribution of the BERTScores for every combination of two questions in the SynthGPT dataset.}
    \includegraphics[width=0.4\textwidth]{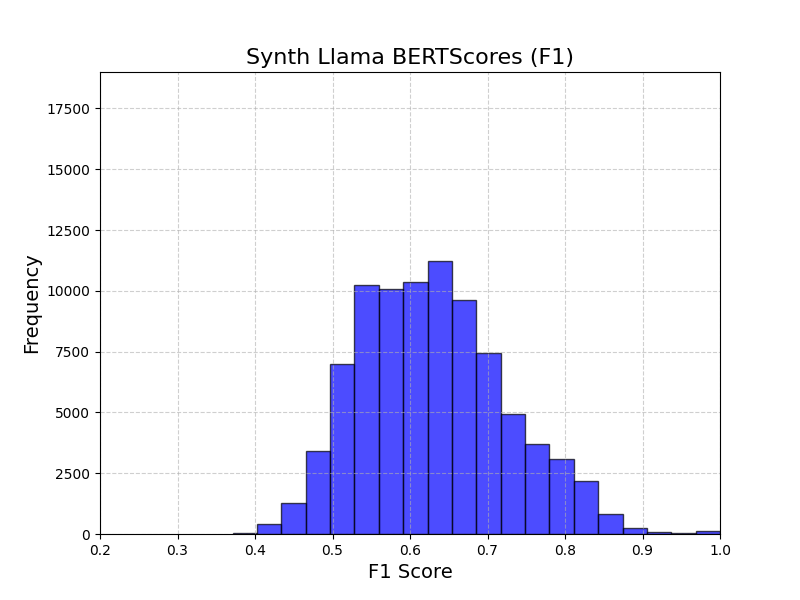}
    \caption{Distribution of the BERTScores for every combination of two questions in the SynthLlama dataset.}
    \label{fig:bertscore_plots}
\end{figure}

Further, we utilize a t-distributed Stochastic Neighbor Embedding (t-SNE) plot to visualize the embedding space of three datasets: human-generated questions, synthetic questions generated by LLaMA, and synthetic questions generated by GPT. The embeddings are extracted from Llama-3-8B-Instruct (the model we finetune in all our experiments), and the t-SNE method reduces the high-dimensional embeddings into a two-dimensional space for visual interpretation.

This visualization allows us to compare the semantic distributions of the datasets and assess how closely the synthetic datasets align with the human-generated questions. Distinct clustering of the datasets in the t-SNE space suggest meaningful differences in their semantic representations. It seems that the two synthetic questions overlap a great deal and have a fair amount of overlap with the crowdsourced questions. However, the crowdsource (human) questions cluster distinctly to the right, outside the space of the synthetic questions. This also suggests greater variety in the crowdsourced questions.

\begin{figure}[H]
    \centering
    \includegraphics[width=0.45\textwidth]{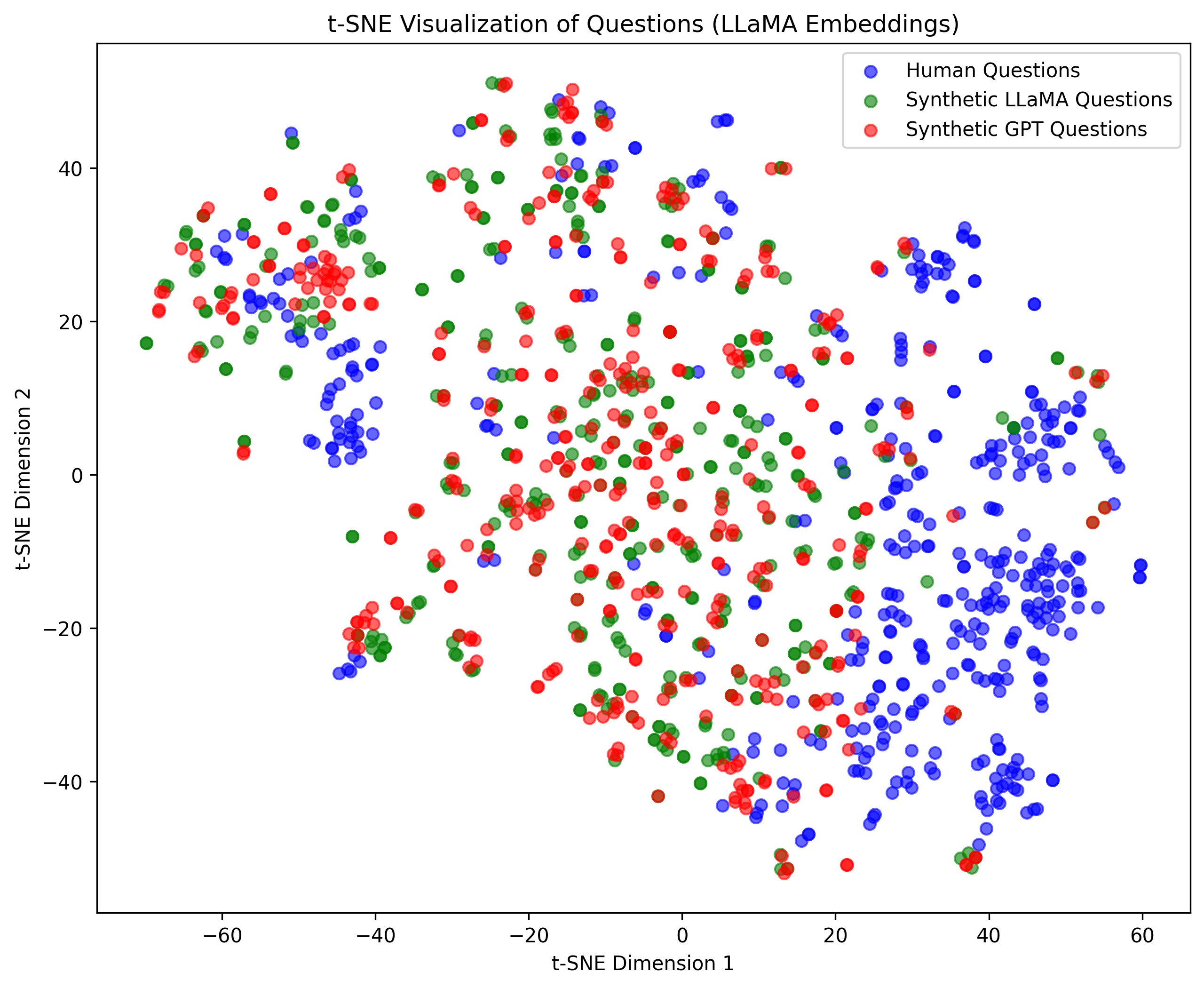}
    \caption{Distribution of the BERTScores for every combination of two questions in the crowdsourced dataset.}
    \label{fig:tSNE}
\end{figure}